# Deep Learning Approaches to Classification of Production Technology for 19$^{th}$ Century Books

Chanjong Im, Junaid Ghauri, John Rothman, Thomas Mandl

University of Hildesheim, Information Science, Germany
mandl@uni-hildesheim.de
imchan@uni-hildesheim.de

**Abstract.** Cultural research is dedicated to understanding the processes of knowledge dissemination and the social and technological practices in the book industry. Research on children books in the 19$^{th}$ century can be supported by computer systems. Specifically, the advances in digital image processing seem to offer great opportunities for analyzing and quantifying the visual components in the books. The production technology for illustrations in books in the 19$^{th}$ century was characterized by a shift from wood or copper engraving to lithography. We report classification experiments which intend to classify images based on the production technology. For a classification task that is also difficult for humans, the classification quality reaches only around 70%.. We analyze some further error sources and identify reasons for the low performance.

## 1 Introduction: Historical Children Book Research

Digital Humanities is having a considerable impact on humanities research related to text. Many text mining tools have been developed and are currently being applied to genuine research questions in the humanities. This trend is currently contributing to a larger variety of methods being used. There has been no comparable paradigm shift in research related to visual material.

Research on historical children books has yet not often been the subject of digital humanities (DH) studies. This research requires processing for both text and images. Children books are considered to contain more images than adult books typically. As a consequence, they are of special interest for an analysis of images. In addition, they form a closed category on the one hand which contains sufficient variety on the other hand.

Illustrated books have played a significant role in knowledge dissemination. The declining production costs for printed images have led to a growing exposure of more and more people to rich visual resources. Research in this area can identify trends in the objects depicted, as well as the stylistic and aesthetic presentation of the visual resources [11]. Images also allow the tracing of influence over time by observing adoptions of content or style.

For cultural studies, it is necessary to observe efficiently whether the shifts in production technology in the 19$^{th}$ century have led to aesthetic developments. Modern technologies like lithography were more adequate for mass production and allowed finer lines.

This shows, that meta data has a great importance for research in children books. However, annotated meta data for production technology is not always available. For many books, the title data can even provide erroneous information on the production technology. Some publishers claimed that they include copper engravings as a marketing argument, however, in many cases this is not true and e.g. lithography images were produced.

As a consequence, one sub task in the DH support for the cultural studies in general and the children books research in particular is the development of robust classification systems which are capable of identifying the production technology based exclusively on visual information.

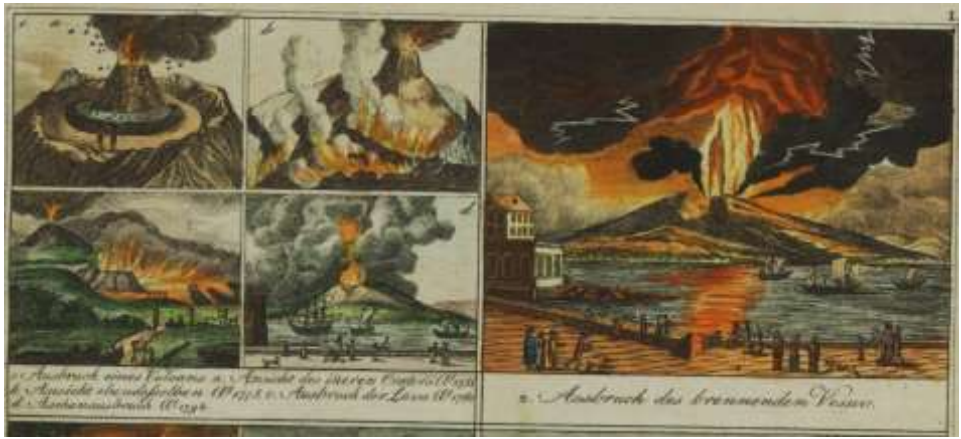

**Fig. 1:** Example of an illustration from the collection

For humans, the identification of the printing technology used for historic books is also not a trivial task. Only experts in the area are able to classify images for this goal. They sometimes need magnifying glasses and the task is often easier when the original paper version is available. For the digital version, this classification is typically even more difficult.

There are also many different forms of lithography (e.g. chromolithography). However, for the purposes of our study the identification of lithography in general is sufficient.

## 2  State of the Art

To the best of the authors' knowledge, this is the first attempt to classify the printing technology using images from books. Other previous work is focused on identifying the positions of text and image blocks within a page. The HBA data challenge for old books intends to improve algorithms for this task [8]. These books are much older than the Hobrecker collection and contain also manuscripts.

Most of the research in image processing is currently being carried out for photographs. These collections vary greatly from the non-realistic drawings and illustrations which can be found in children books. It is unclear how CNN and other models optimized for photographs perform for these images.

Also in the analysis of art, there is little work on mining approaches. Few researchers have processed large amounts of images in this field. One experiment by Salah & Elgammel is dedicated to classify the painter of artistic work. Such work is highly dependent on the type of paintings in the collection [10].

A recent project is focusing on graphic novels. Current state of the art CNNs are applied to tasks like author identification with very good success [6]. In addition, the processing is aimed at measuring the drawing style of a graphic novel in order to find similar books. However, the production technology is not relevant for these modern books.

The work that seems to be most similar to the classification of historical printing technologies is research in the area of texture recognition. Some authors manage to classify material from visual data. The materials are different types of plastic. The algorithm learns the typical texture structure of each material based on magnified data. For a material classification task, narrow structures for a CNN were used [1]. There has also been work towards applying deep networks and transfer learning for material classification. Cimpoi et al. conducted material classification with deeper structure and transfer learning [4]. Some authors achieved very successful results showing an accuracy of 99 to 99.9 percent with specially constructed dataset i.e. CU-ReT [5].

## 3  Data Collection and Classification

One of the first goals for the research of images in historical children books lies within the production technologies. As a classification problem with few classes, it seems like a challenge which could be solved with current technology.

Our primary assumption is, that differences in printing technology can be observed only at the detail level. Therefore, resizing of the images is being avoided. Much rather, the decision was made to extract small parts of the images in full resolution in order to retain the minute details of the printing. It is much less relevant to size images and make sure all objects are fully contained in an image because the problem seems not related to object identification.

Consequently, the few available training images were cropped into small parts without re-sizing them which also leads to more training examples.

### 3.1 Data Collection

The data collection is based on the Hobrecker collection. Karl Hobrecker was a collector of children books. His books are now archived in the library of the Technical University of Braunschweig. A subset has been digitized and is available online [13]. The collection is of great interest for cultural research. It contains different types of children books mainly from the 19$^{th}$ century: e.g. alphabetization books, biographies, natural history descriptions as well as adventure and travel stories. The scanned Hobrecker collection of children's books available for digital analysis contains around 350 books. There is no knowledge about the distribution of printing technologies within the entire set. It would require too much human effort to manually classify all books.

Out of these, 32 books were carefully labeled by experts and are used for the classification tasks. The experts classified the books into either wood engraving or lithography.

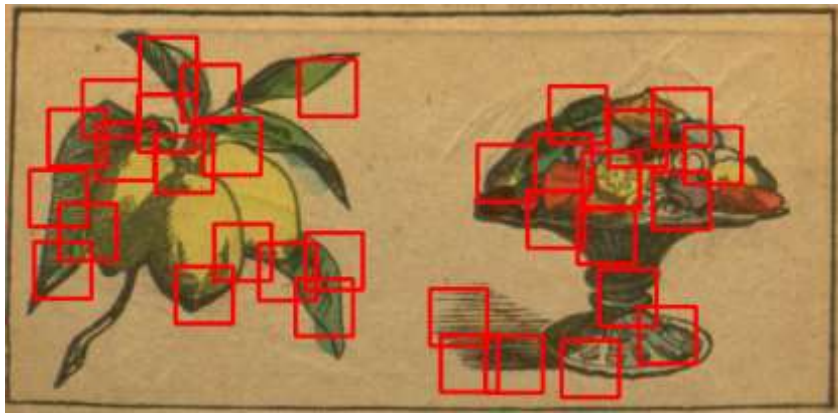

**Fig. 2:** Examples cropping using slicing and a contour method

The images in the scanned books were separated from the text by using image preprocessing techniques as described by Ban [2]. To minimize the pixel information loss on images, all the images are cropped into the size of 128 * 128 pixels. In order to avoid an imbalance problem for classification, the number of crops are set equally by uniformly sampling from each book. An example for a crop is shown in figure 2.

Table 1 shows some general statistics of the 32 books sorted by the two printing technologies.

Table 1. Training set statistics.

| Printing type | Number of books | Number of images | Number of crops |
|---|---|---|---|
| Wood Engraving | 14 | 349 | 2235 |
| Lithography | 18 | 173 | 2235 |

### 3.2 Classification

Convolution Neural Network (CNN), the recent state of the art technology is known to be very effective in automated feature detection and subsequent classification in many domains [e.g. 14,15]. In the approach presented in this paper, CNNs are used as the processing model for the printing type classification.

Similar tasks have conducted material classification with a narrow structure of CNN. Mehri et al. conducted material classification with deeper structure and transfer learning [8].

The major difference between the dataset used in material classification and the Hobrecker dataset is that the Hobrecker dataset contains a lot more noisy features due to the nature of scanned images from old books which are often compromised by age. The dataset used for material classification is created for the main purpose of material analysis, whereas the material for the books is always paper. However the quality and type differ greatly.

Two methods are used for the task. One is using the pre-trained model. We applied the Inception Network as suggested by Szegedy [12], for feature extraction and use these features to feed fully connected neural networks (FCN) as well as Support Vector Machine (SVM) for classification. The second method uses a slim CNN architecture, similar to the one used by Ban [2], and trains the model with a randomized weight initialization.. Several modifications are made on top of this model. Firstly, the input size is increased from 64 *64 to 128 * 128. Secondly, two kinds of deeper networks with more convolution layers and pooling layers are tested. The architectures are shown in Table 2.

These architectures were used for the following motivation. The size the of the features space to be learned is unclear. So we initially selected two different filter sizes to account for that.

For the smaller filter size, two different layer structures were applied. The first one uses one pooling layer for each convolution layer. The second one uses fewer pooling layer.

**Table 2.** Two CNN architectures used for printing type classification. Values in the brackets of Conv represents the kernel size. For example, Conv(11) is convolution layer with (11*11) sized kernel. * represents number of same layers in the flow. Pool represents average pooling with (2*2) sized kernel and stride of 2.

| Name | Architecture |
|---|---|
| Big-Filters | Conv(11)-Pool-Conv(10)-Pool-Conv(6)-Pool-Conv(3)-Pool-Conv(3)-Pool |
| Small-Filters-Less Pooling | Conv(3)*5-Pool-Conv(3)*4-Conv(2)-Pool-Conv(6)-Pool-Conv(3)-Pool-Conv(3)-Pool |
| Small-Filters-Balanced Pooling | Conv(3)-Pool-Conv(4)-Pool-(Conv(3)-Pool)*3-Conv(2) |

## 4 Results and analysis

### 4.1 Results

Unlike the results of CUReT [5], transfer learning results turned out to be poor of maximum of 58 % of accuracy. This accuracy is achieved when using FCN of two hidden layers with 512 nodes respectively using the features passed through by the pre-trained Inception model. Replacing FCN with SVM decreased the performance leading to 47% for linear kernel and 52% for non-linear kernel.

The results using the model from Ban [2] showed 47% of accuracy and a similar model adopted from Mehri et al. [8] showed 51% of accuracy at the most. The results of the small-filters less-pooling architecture achieved 61% accuracy and Big-Filters showed the best performance of 63% classification accuracy. Small-Filters-Balanced-Pooling showed the lowest performance of 48% among the architectures which were tested. An overview of the results can be seen in Table 3.

**Table 3.** Results of the CNN architectures

| Name | Accuracy |
|---|---|
| Big-Filters | 63% |
| Inception Model with Neural Networks | 58% |
| Inception Model with linear SVM | 47% |
| Inception Model with linear SVM | 52% |
| Small-Filters-Less Pooling | 61% |
| Small-Filters-Balanced Pooling | 48% |

Some examples for misclassified iages are hown in Figure 3.

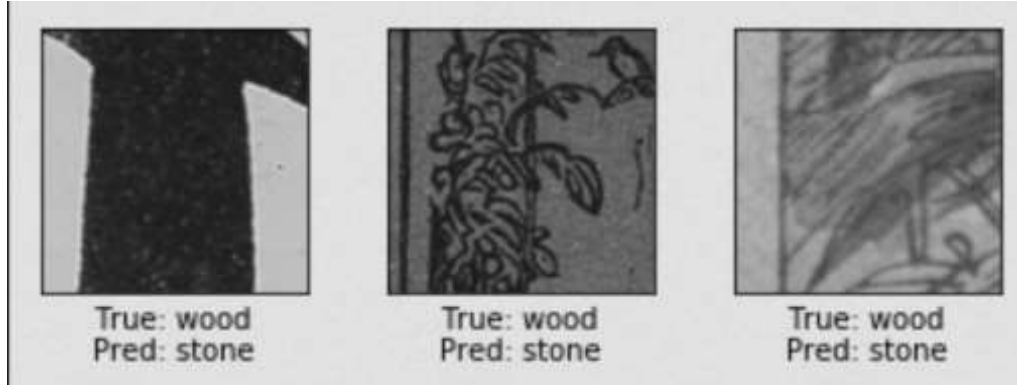

**Fig. 3.** Three misclassified examples.

### 4.2 Analysis and Discussion

There are three potential reason for the low accuracy we obtained in the results.

- Noisy dataset: The nature of scanned files leads to difficulties in automated image extraction and preprocessing. For example, in the image placed left in fig.2, a part of the character is cropped from the scanned file. Such kind of crops are not eliminated in the dataset due to background noise confusing the algorithms used in extraction phase. Moreover, quality of paper that is scanned is not in equal conditions across books. For example, some books tend to be in a very good shapes with no sign of text marks printed on top of image. CNNs react very sensitive to this noise when small filters are used along with pooling methods. Using bigger sized filters at the initial stage of convolution operations, gives a model a flexibility on the noises.
- The original assumption about crops might not be correct. Rather full knowledge of an image might be required to classify. This seems to be also relevant for humans. Some of the misclassified cropped images were presented to experts and they could not easily distinguish between the printing technologies. They required to see the full image. For future experiments, it is intended to work with full images. It seems the experts need to see minute details as shown in crops but also the flow of lines overall in order to reach a definite decision. As a consequence, future classification experiments will consider the output of high and low levels within the CNN.
- Insufficient amount of data: To perform tasks with deep learning, more data is needed for training.

## 5  Outlook

For optimizing the identification of the printing type, more data is obtained. Labelled images from the Pictura Paedagogica Online (PPO) [3] are obtained and will be processed.

Currently, experiments with object recognition on the collection are being carried out with Yolo [9] which is a pre-trained model based on photographs. It is interesting to check whether it also works for non-realistic images as we find them in historic books. First results indicate that the performance is very different for individual classes. The object recognition with a satisfying quality would allow the humanities scholars to trace the frequency of objects and the introduction of new knowledge to children's books.

## Acknowledgements

We thank the Fritz Thyssen Foundation for their funding for the research project *Distant Viewing*. We would like to thank the library of the Technische Universität Braunschweig for facilitating access to the digitized Hobrecker collection.